\documentclass[10pt,twocolumn,letterpaper]{article}

\usepackage{cvpr}
\usepackage{times}
\usepackage{epsfig}
\usepackage{graphicx}
\usepackage{amsmath}
\usepackage{amssymb}
\usepackage{subfig}
\usepackage{float}

\usepackage[breaklinks=true,bookmarks=false]{hyperref}

\cvprfinalcopy 

\def\cvprPaperID{9557} 

\ifcvprfinal\fi
\begin{document}

\title{Learning a Unified Sample Weighting Network for Object Detection\thanks{{\small This work was performed at JD AI Research.}}}

\author{Qi Cai$^{\dag}$, Yingwei Pan$^{\ddag}$, Yu Wang$^{\ddag}$, Jingen Liu$^{\S}$, Ting Yao$^{\ddag}$, and Tao Mei$^{\ddag}$ \\
{\small\centering$^{\dag}$ University of Science and Technology of China, Hefei, China}\\
{\small\centering$^{\ddag}$ JD AI Research, Beijing, China}~~~
{\small\centering$^{\S}$ JD AI Research, Mountain View, USA}\\
{\tt\scriptsize \{cqcaiqi, panyw.ustc, feather1014, jingenliu, tingyao.ustc\}@gmail.com, tmei@jd.com}
}

\maketitle
\thispagestyle{empty}

\begin{abstract}
	Region sampling or weighting is significantly important to the success of modern region-based object detectors. Unlike some previous works, which only focus on ``hard'' samples when optimizing the objective function, we argue that sample weighting should be data-dependent and task-dependent. The importance of a sample for the objective function optimization is determined by its uncertainties to both object classification and bounding box regression tasks. To this end, we devise a general loss function to cover most region-based object detectors with various sampling strategies, and then based on it we propose a unified sample weighting network to predict a sample's task weights. Our framework is simple yet effective. It leverages the samples' uncertainty distributions on classification loss, regression loss, IoU, and probability score, to predict sample weights. Our approach has several advantages: (i). It jointly learns sample weights for both classification and regression tasks, which differentiates it from most previous work. (ii). It is a data-driven process, so it avoids some manual parameter tuning. (iii). It can be effortlessly plugged into most object detectors and achieves noticeable performance improvements without affecting their inference time. Our approach has been thoroughly evaluated with recent object detection frameworks and it can consistently boost the detection accuracy. Code has been made available at \url{https://github.com/caiqi/sample-weighting-network}.
\end{abstract}
\section{Introduction}

Modern region-based object detection is a multi-task learning problem, which consists of object classification and localization. It involves region sampling (sliding window or region proposal), region classification and regression, and non-maximum suppression. Leveraging region sampling, it converts object detection into a classification task, where a vast number of regions are classified and regressed. According to the way of region search, these detectors can be categorized into one-stage detectors \cite{lin2017focal, liu2016ssd, redmon2016you,zhang2018single} and two-stage detectors \cite{cai2018cascade, girshick2015fast, girshick2014rich,he2017mask,lin2017feature,ren2015faster}.

\begin{figure}[!tb]
	\vspace{-0.12in}
	\centering {\includegraphics[width=0.43\textwidth]{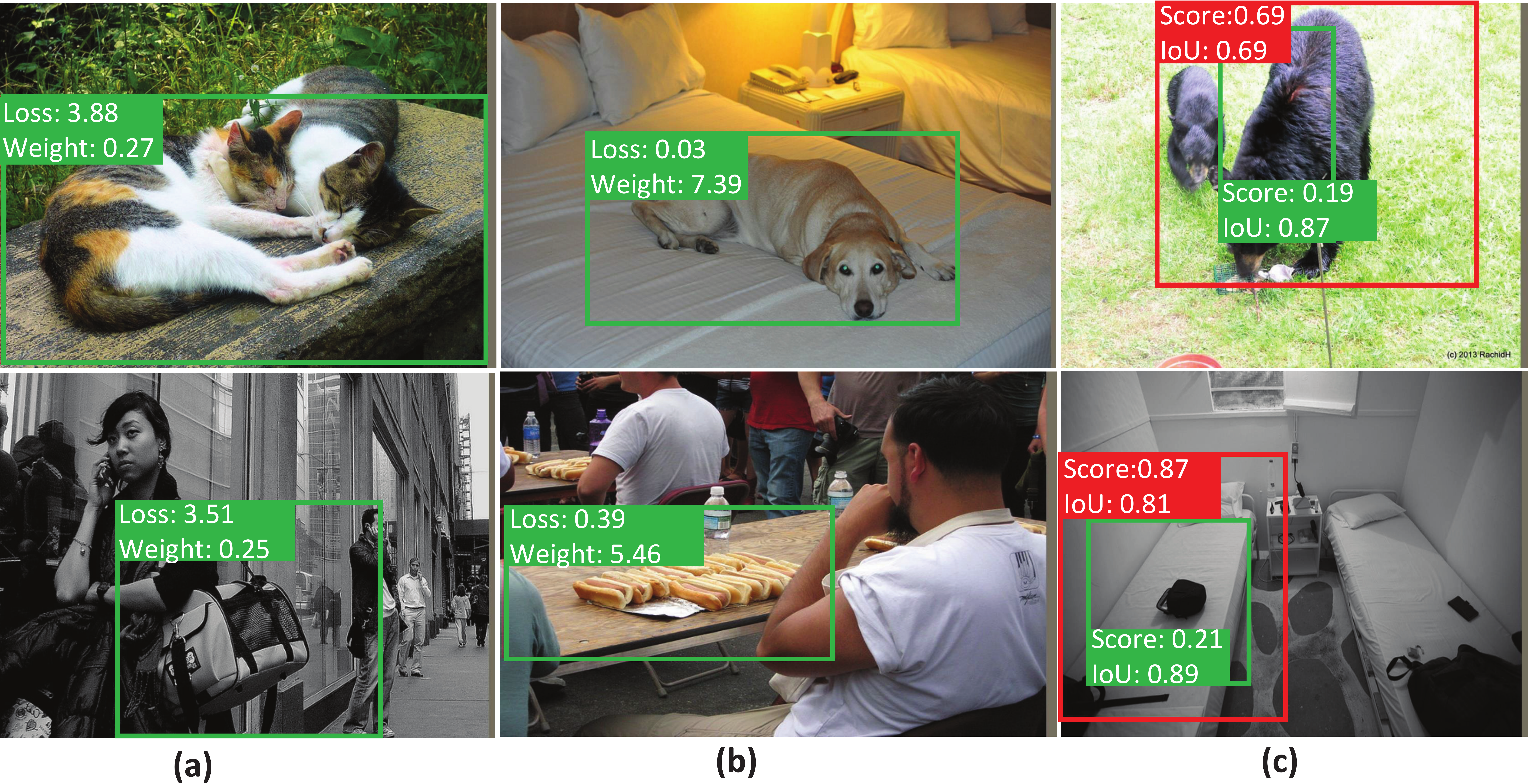}}
	\vspace{-0.12in}
	\caption{\small Samples from our training process. (a) The samples having large classification loss but small weight. (b) The samples having small classification loss but large weight. (c) The samples exhibiting inconsistency between classification score and IoU.}
	\label{fig:intro_sample}
	\vspace{-0.22in}
\end{figure}

In general, the object detectors of the highest accuracy are based on the two-stage framework such as Faster R-CNN \cite{ren2015faster}, which rapidly narrows down regions (most of them are from the background) during the region proposal stage. In contrast, the one-stage detectors, such as SSD \cite{liu2016ssd} and YOLO \cite{redmon2016you}, achieve faster detection speed but lower accuracy. It is because of the class imbalance problem (i.e., the imbalance between foreground and background regions), which is a classic challenge for object detection.

The two-stage detectors deal with class imbalance by a region-proposal mechanism followed by various efficient sample strategies, such as sampling with a fixed foreground-to-background ratio and hard example mining \cite{felzenszwalb2010DPM,shrivastava2016training,viola2001adaboost}. Although the similar hard example mining can be applied to one-stage detectors, it is inefficient due to a large number of easy negative examples \cite{lin2017focal}. Unlike the Online Hard Example Mining (OHEM) \cite{shrivastava2016training} which explicitly selects samples with high classification losses into the training loop, Focal-Loss \cite{lin2017focal} proposes a soft weighting strategy, which reshapes the classification loss to automatically down-weight the contributions of easy samples and thus focuses the training on hard samples. As a result, the manually tuned Focal-Loss can significantly improve the performance of one-stage detectors.

The aforementioned ``hard'' samples generally refer to those with large classification loss. However, a ``hard'' sample is not necessarily important. As Figure~\ref{fig:intro_sample} (a) (All samples are selected from our training process.) illustrates, the samples have high classification losses but small weights (``hard'' but not important). Conversely, an ``easy'' sample can be significant if it captures the gist of the object class as shown in Figure~\ref{fig:intro_sample} (b). In addition, the assumption that the bounding box regression is accurate when the classification score is high, does not always hold as examples shown in Figure~\ref{fig:intro_sample} (c). There may be a misalignment between classification and regression sometimes \cite{jiang2018acquisition}. Hence, an IoU-Net is proposed in \cite{jiang2018acquisition} to predict a location confidence. Furthermore, there are ambiguities in bounding box annotations due to occlusion, inaccurate labeling, and ambiguous object boundary. In other words, the training data has uncertainties. Accordingly, \cite{he2019bounding} proposes a KL-Loss to learn bounding box regression and location uncertainties simultaneously. The samples with high uncertainty (high regression loss) are down-weighted during training.

Sample weighting is a very complicated and dynamic process. There are various uncertainties, which exist in individual samples when applying to a loss function of a multi-task problem. Inspired by \cite{kendall2018multi}, we argue that sample weighting should be \textit{data-dependent} and \textit{task-dependent}. On the one hand, unlike previous work, the importance of a sample should be determined by its intrinsic property compared to the ground truth and its response to the loss function. On the other hand, object detection is a multi-task problem. A sample's weights should balance among different tasks. If the detector trades its capacity for accurate classification and generates poor localization results, the mislocalized detection will harm average precision especially under high IoU criterion and vice versa.

Following the above idea, we propose a unified dynamic sample weighting network for object detection. It is a simple yet effective approach to learn sample-wise weights, which also balances between the tasks of classification and regression. Specifically, beyond the base detection network, we devise a sample weighting network to predict a sample's classification and regression weights. The network takes classification loss, regression loss, IoU and score as inputs. It serves as a function to transform a sample's current contextual feature into sample weight. Our sample weighting network has been thoroughly evaluated on MS COCO \cite{lin2014microsoft} and Pascal VOC \cite{everingham2010pascal} datasets with various one-stage and two-stage detectors. Significant performance gains up to 1.8\% have been consistently achieved by ResNet-50 \cite{he2016resnet} as well as a strong ResNeXt-101-32x4d \cite{xie2017aggregated} backbone. The ablation studies and analysis further verify the effectiveness of our network and unveil its internal process.

In summary, we propose a general loss function for object detection, which covers most region-based object detectors and their sampling strategies, and based on it we devise a unified sample weighting network. Compared to previous sample weighting approaches \cite{cao2019prime, girshick2014rich,he2019bounding,lin2017focal}, our approach has the following advantages: (i). It jointly learns sample weights for both classification task and regression task. (ii). It is data-dependent, which enables to learn soft weights for each individual sample from the training data. (iii). It can be plugged into most object detectors effortlessly and achieves noticeable performance gains without affecting the inference time.

\section{Related Work}
\textbf{Region-based object detection} can be mainly categorized into two-stage and one-stage approaches. The two-stage approaches, e.g., R-CNN \cite{girshick2014rich}, Fast R-CNN \cite{girshick2015fast} and Faster R-CNN \cite{ren2015faster}, consist of region proposal stage and region classification stage. Various region proposal techniques have been devised, such as selective search \cite{uijlings2013selective} and Region Proposal Network \cite{ren2015faster}. In the second stage, regions are classified into object categories and bounding box regression is performed simultaneously. Significant improvements have been made by new designed backbones \cite{chen2019detnas,dai2017deformable,lin2017feature}, architectures \cite{cai2018cascade,chen2019hybrid,dai2016r}, and individual building blocks \cite{deng2019relation,hu2018relation,jiang2018acquisition,lu2019grid,wang2019region}. Inspired by domain adaptation for recognition \cite{pan2019transferrable,yao2015semi}, another line of research \cite{cai2019exploring,chen2018domain,khodabandeh2019robust} focuses on learning robust and domain-invariant detectors based on two-stage approaches. In contrast, one-stage approaches including SSD \cite{liu2016ssd} and YOLO \cite{redmon2016you} remove the region proposal stage and directly predict object categories and bounding box offsets. This simplicity gains faster speed at the cost of degradation of accuracy.

Our sample weighting network (SWN) is devised to boost general region-based object detectors. It can be easily plugged into the aforementioned object detectors without adding much training cost. In fact, it does not affect the inference at all, which makes our approach very practical.

\textbf{ Region sampling or weighting strategy} plays an important role in the training of object detection models. Random sampling along with a fixed foreground-background ratio is the most popular sampling strategy for early object detection \cite{girshick2015fast,ren2015faster}. However, not every sample plays equal importance to optimization. Actually, the majority of negative samples are easy to be classified. As a result, various hard example mining strategies have been proposed, including hard negative examples mining \cite{girshick2014rich,liu2016ssd}, Online Hard Example Mining (OHEM) \cite{shrivastava2016training}, and IoU guided sampling \cite{cai2018cascade,lin2017focal}. Instead of making hard selection, Focal-Loss \cite{lin2017focal} proposes to assign soft-weights to samples, such that it reshapes the classification loss to down-weight ``easy'' samples and focus training on ``hard'' ones. However, some recent works \cite{cao2019prime,wu2019iou} notice ``easy'' samples may be also important. Prime sampling \cite{cao2019prime} and IoU-balanced loss \cite{wu2019iou} have been advanced to make ``easy'' samples more important for loss function optimization.

Beyond various sample weighting approaches, we devise a general loss function formulation which represents most region-based object detectors with their various sampling strategies. Based on this formulation, we design a unified sample weighting network to adaptively learn individual sample weights. Rather than manually crafted based on certain heuristics \cite{cao2019prime, lin2017focal}, our sample weighting network is directly learned from the training data. In addition, unlike most existing methods \cite{he2019bounding,lin2017focal} designed for classification or regression, our approach is able to balance the weights between the classification and regression tasks.

\textbf{Multi-task sample weighting} has two typical directions of function design. One capitalizes on a monotonically increasing function w.r.t. the loss value, such as AdaBoost \cite{freund1997decision} and hard example mining \cite{malisiewicz2011ensemble}. The other designs monotonically decreasing function w.r.t. the loss value, especially when the training data is noisy. For example, Generalized Cross-Entropy \cite{zhang2018generalized} and SPL \cite{kumar2010self} propose to focus more on easy samples. Recently, some learning-based approaches are proposed to adaptively learn weighting schemes from data, which eases the difficulty of manually tuning the weighting function \cite{fan2018learning,jiang2017mentornet,ren2018learning, shu2019meta}. In the regime of multi-task learning, \cite{kendall2018multi} proposes using homoscedastic task uncertainty to balance the weighting between several tasks optimally where the tasks with higher uncertainties are down-weighted during training.

\section{A Unified Sample Weighting Network}

\subsection{Review of Sampling Strategies}
In this section, we briefly review the training objectives and sampling strategies for object detection. Recent research on object detection including one-stage and two-stage object detectors follows a similar region-based paradigm. Given a group of anchors $a_i \in \mathcal{A}$, i.e., prior boxes, which are regularly placed on an image to densely cover spatial positions, scales and aspect ratios, we can summarize the multi-task training objective as follows:
\setlength{\belowdisplayskip}{2pt} \setlength{\belowdisplayshortskip}{2pt}
\setlength{\abovedisplayskip}{2pt} \setlength{\abovedisplayshortskip}{2pt}
\begin{equation}\label{eq:loss_1}
	L = \frac{1}{N_1} \sum_{ \{i:a_i \in \mathcal{A}^{cls}\}}L_{i}^{cls} + \frac{1}{N_2} \sum_{ \{i:a_i \in \mathcal{A}^{reg}\}} L_{i}^{reg},
\end{equation}
where $L_{i}^{cls}$ ($L_{i}^{reg}$) is the classification loss (regression loss), and $\mathcal{A}^{cls}$ ($\mathcal{A}^{reg}$) denotes the sampled anchors for classification (regression). $N_1$ and $N_2$ are the number of training samples and foreground samples. The relation $\mathcal{A}^{reg} \subset \mathcal{A}^{cls} \subset \mathcal{A}$ holds for most object detectors. Now, let $s_{i}^{cls}$ and $s_{i}^{reg}$ be sample $a_i$'s weights for the classification and regression losses respectively, we formulate a generalized loss function for both two-stage and one-stage detectors with various sampling strategies, by converting Eq.~\ref{eq:loss_1} to:
\setlength{\belowdisplayskip}{2pt} \setlength{\belowdisplayshortskip}{2pt}
\setlength{\abovedisplayskip}{2pt} \setlength{\abovedisplayshortskip}{2pt}
\begin{equation}\label{eq:loss_2}
	L = \frac{1}{N_1} \sum_{ \{i:a_i \in \mathcal{A}\} } s_i^{cls} L_{i}^{cls} + \frac{1}{N_2} \sum_{ \{i:a_i \in \mathcal{A}\} } s_i^{reg} L_{i}^{reg},
\end{equation}
where $s_i^{cls}=\emph{I}[a_i \in \mathcal{A}^{cls}]$ and $s_i^{reg}=\emph{I}[a_i \in \mathcal{A}^{reg}]$. $\emph{I}[\cdot]$ is the indicator function which outputs one when condition satisfied, otherwise zero. As a result, we can employ $S^{cls}=\{s_i^{cls}\}$ and $S^{reg}=\{s_i^{reg}\}$ to represent various existing sample strategies. Here, we reinterpret region sampling as a special case of sample weighting, which allows for soft sampling. In the following paragraph, we will briefly explain most popular sampling or weighting approaches under our general loss formulation.

\subsection{Problems in Existing Sampling Approaches}
\begin{figure}[!tb]
	\centering {\includegraphics[width=0.45\textwidth]{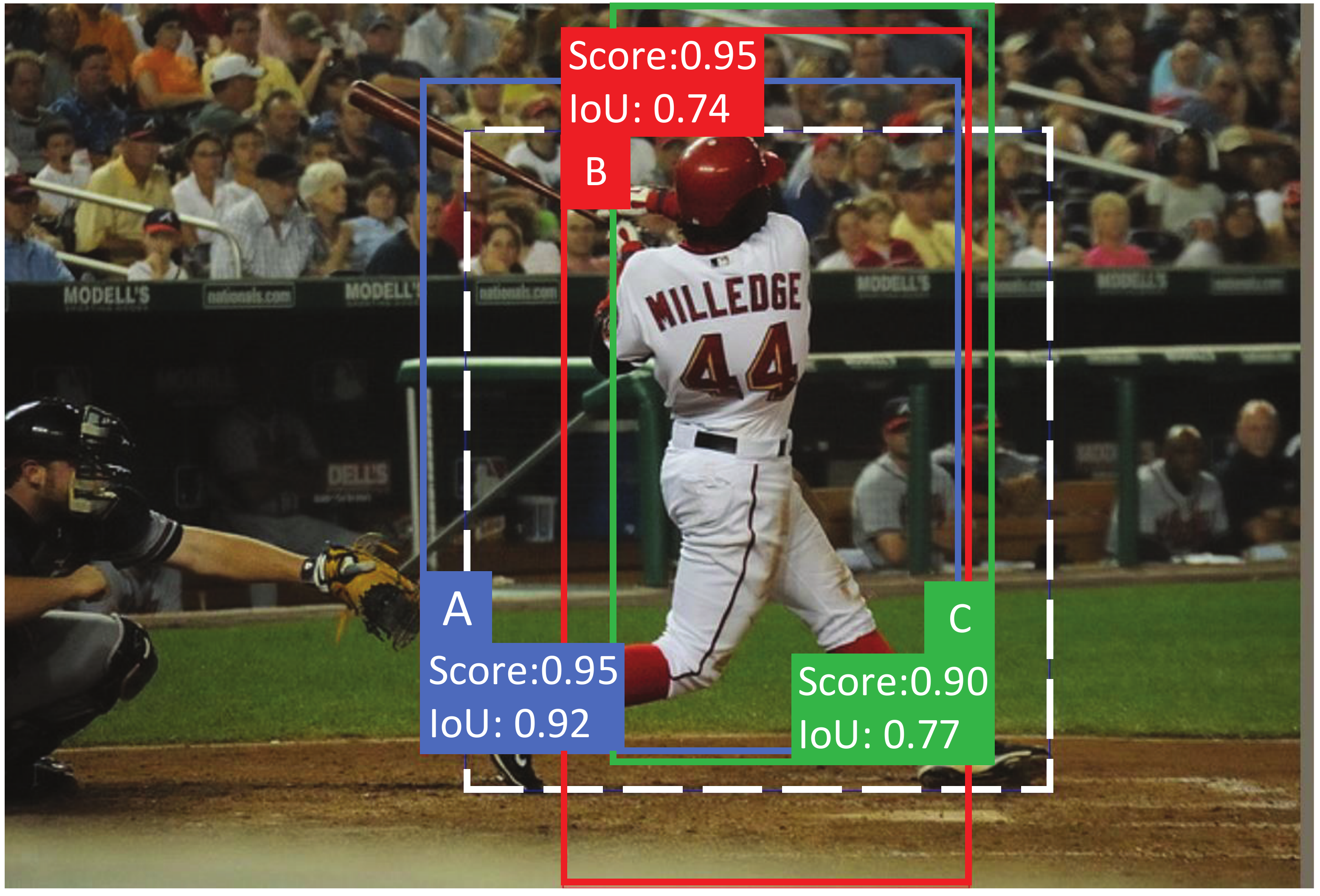}}
	\vspace{-0.1in}
	\caption{\small Training samples of Faster R-CNN after the first epoch. The dashed white box denotes the ground truth. $A, B, C$ are three positive samples with different predicted scores and IoUs.}
	\label{fig:tech_sample}
	\vspace{-0.18in}
\end{figure}

\textbf{RPN, Random Sampling and OHEM}
Region Proposal Network (RPN) classifies each sample into class-agnostic foreground or background class. Taking RPN as a data-driven sampling strategy, the classification weight for $a_i$ is defined as: $s_{i}^{cls}=\emph{I}[p (a_i) > \rho]*\emph{I}[a_i \in \mathcal{A}_{NMS}]$ where $\rho$ is the threshold to filter out samples with low foreground scores, and $\mathcal{A}_{NMS}$ is the anchor set after applying Non-Maximum-Suppression (NMS). Random Sampling uniformly selects $n_p$ samples from $\mathcal{A}^P$ (positive) and $n_n$ samples from $\mathcal{A}^N$ (negative), where $n_p$ and $n_n$ represent the required number of positive and negative samples, respectively. The classification weights for selected samples are assigned to be 1, while the rest to be 0. Instead of randomly sampling with equal probability, OHEM first ranks positive and negative samples separately in a monotonically decreasing order based on their loss values. Then the classification weights of top-$n_p$ positive and top-$n_n$ negative samples are assigned to be 1, and the rest to be 0. For all sampling, their samples' regression weights can be defined as $s_{i}^{reg}=\emph{I}[s_{i}^{cls} = 1]*\emph{I}[a_i \in \mathcal{A}^P]$.

\textbf{Focal-Loss and KL-Loss}
Focal-Loss reshapes the loss function to down-weight easy samples and focus the training on hard ones. It can be regarded as assigning soft classification weight to each sample: $s_{i}^{cls} = (1 - p (a_i)) ^\gamma$ where $\gamma >0$. And the regression loss are computed on all positive samples, $s_{i}^{reg}=\emph{I}[a_i \in \mathcal{A}^P]$. KL-Loss re-weights the regression loss depending on the estimated uncertainty $\sigma_i^2$: $s_{i}^{reg} = 1 / \sigma_i^2$. The classification weights are the same as that of Random Sampling and OHEM.

Given a set of anchors $\mathcal{A} = \mathcal{A}^P \cup \mathcal{A}^N$, the goal of sample weighting is to find a weighting assignments $S^{cls}$ and $S^{reg}$ for better detection performance. Now, let us have a close inspection of two important components, i.e., NMS and mAP, to understand their particular roles in sample weighting. In general, the NMS filters cluttered bounding boxes by removing the boxes having relatively low scores. Taking the three boxes $A$, $B$, $C$ in Figure~\ref{fig:tech_sample} for example, $C$ is suppressed during the inference due to its relatively lower score compared with $A$ and $B$. In contrast, when OHEM is applied, $C$ will be selected for training due to its higher loss (lower score). Putting too much attention to ``hard'' examples like ``C'' may not be always helpful, because during the inference we also pursue a good ranking. Focal-Loss also faces a similar problem as it assigns the same classification weight to box $A$ and $B$. But, given that the IoU of $A$ with regard to the ground truth is higher than that of $B$, aiming at improving the score of A would potentially be more beneficial. This is because the mAP is computed at various IoU thresholds, which favors more precisely localized detection results. KL-Loss, on the other hand, assigns different sample weights for regression loss based on bounding box uncertainty, while it ignores re-weighting classification loss.

Given these drawbacks of existing methods, we propose to learn sample weights jointly for both classification and regression from a data-driven perspective. Briefly speaking, previous methods concentrate on re-weighting classification (e.g., OHEM \& Focal-Loss) or regression loss (e.g., KL-Loss). But our approach jointly re-weights classification and regression loss. In addition, being different from mining ``hard'' examples in OHEM \& Focal-Loss approaches, which have higher classification loss, our approach focuses on important samples, which could be ``easy'' ones as well.

\subsection{Joint Learning for Sample Weighting}
Inspired by a recent work on uncertainty prediction for multi-task learning \cite{kendall2018multi}, we reformulate the sample weighting problem in a probabilistic format, and measure the sample importance via the reflection of uncertainties. We demonstrate that our proposed method enables the sample weighting procedure to be flexible and learnable via deep learning. Note that our work differentiates from \cite{kendall2018multi}, because our probabilistic modeling addresses not only the sample wise weighting, but also the balance between classification and localization tasks. Yet, the work \cite{kendall2018multi} only considers the multi-task setting where all training samples share the same weights.

\begin{figure*}[!tb]
	\vspace{-0.2in}
	\centering {\includegraphics[width=1.0\textwidth]{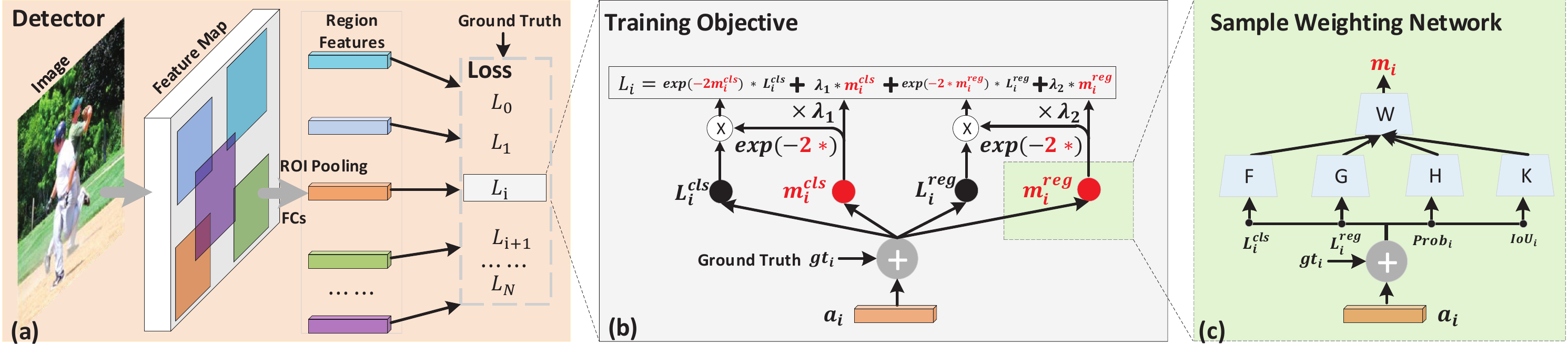}}
	\vspace{-0.2in}
	\caption{\small The framework for our Sample Weighting Network (SWN). (a) The general framework for a two-stage detector (it can be replaced with one-stage detector). In the forward pass, each sample is compared with its ground truth. The classification and regression losses are computed. In the backward pass, the loss of all samples are averaged to optimize the model parameters. (b) The break down of loss function which supervises the base detection network and SWN. The gradient can be backpropagated to the detection network and sample weighting network. (c) depicts the SWN design. It absorbs $L_i^{cls}$, $L_i^{reg}$,$Prob_i$,$IoU_i$ as input and generates weights for each sample.}
	\label{fig:framework}
	\vspace{-0.2in}
\end{figure*}

The object detection objective can be decomposed into regression and classification tasks. Given the $i^{\text{th}}$ sample, we start by modeling the regression task as a Gaussian likelihood, with the predicted location offsets as mean and a standard deviation $\sigma_{i}^{reg}$:
\setlength{\belowdisplayskip}{2pt} \setlength{\belowdisplayshortskip}{2pt}
\setlength{\abovedisplayskip}{2pt} \setlength{\abovedisplayshortskip}{2pt}
\begin{equation}\label{eq:loss_3}
	p (gt_{i}|a^*_i) = \mathcal{N} (a^*_i, {\sigma^{reg}_i}^2),
\end{equation}
where vector $gt_i$ represents the ground truth bounding box coordinates, and $a^*_i$ is the estimated bounding box coordinates. In order to optimize the regression network, we maximize the log probability of likelihood:
\setlength{\belowdisplayskip}{2pt} \setlength{\belowdisplayshortskip}{2pt}
\setlength{\abovedisplayskip}{2pt} \setlength{\abovedisplayshortskip}{2pt}
\begin{equation}\label{eq:loss_4}
	\log p (gt_i|a^*_i) \propto - \frac{1}{{\sigma^{reg}_i}^2} ||gt_i -a^*_i||_{2}^2 - \log\sigma^{reg}_i,
\end{equation}
By defining $L_{i}^{reg}=||gt_i -a^*_i||_{2}^2$, multiplying Eq.~\ref{eq:loss_4} with $-1$ and ignoring the constant, we obtain the regression loss:
\setlength{\belowdisplayskip}{2pt} \setlength{\belowdisplayshortskip}{2pt}
\setlength{\abovedisplayskip}{2pt} \setlength{\abovedisplayshortskip}{2pt}
\begin{equation}\label{eq:loss_5}
	L_i^{reg*} = \frac{1}{{\sigma^{reg}_i}^2} L_{i}^{reg} + \lambda_2 \log\sigma^{reg}_i,
\end{equation}
where $\lambda_2$ is a constant value absorbing the global loss scale in detection objective. By writing $ 1/ {\sigma^{reg}_i}^2$ as $s_i^{reg}$, Eq.~\ref{eq:loss_5} can be roughly viewed as a weighted regression loss with a regularization term preventing the loss from reaching trivial solutions. As the deviation increases, the weight on $L_{i}^{reg}$ decreases. Intuitively, such weighting strategy places more weights on confident samples and penalizes more on mistakes made by these samples during training. For classification, the likelihood is formulated as a softmax function:
\setlength{\belowdisplayskip}{2pt} \setlength{\belowdisplayshortskip}{2pt}
\setlength{\abovedisplayskip}{2pt} \setlength{\abovedisplayshortskip}{2pt}
\begin{equation}\label{eq:loss_6}
	p (y_i|a^*_i) = softmax (y_i, \frac{1}{t_i} p (a^*_i)),
\end{equation}
where the temperature $t_i$ controls the flatness of the distribution. $p (a_i^*)$ and $y_i$ are the unnormed predicted logits and ground truth label of $a_i^*$, respectively. The distribution of $p (y_i|a^*_i) $ is in fact a Boltzmann distribution. To make its form consistent with that of the regression task, we define $t_i$ = 1/${\sigma_i^{cls}}^2$. Let $L_{i}^{cls} =-\log {softmax} (y_i,p (a^*_i))$, the classification loss is approximated by:
\setlength{\belowdisplayskip}{2pt} \setlength{\belowdisplayshortskip}{2pt}
\setlength{\abovedisplayskip}{2pt} \setlength{\abovedisplayshortskip}{2pt}
\begin{equation}\label{eq:loss_7}
	L_i^{cls*} = \frac{1}{{\sigma^{cls}_i}^2} L_{i}^{cls} + \lambda_1 \log\sigma^{cls}_i,
\end{equation}
Combining weighted classification loss Eq.~\ref{eq:loss_7} and weighted regression loss Eq.~\ref{eq:loss_5} yields the overall loss:
\setlength{\belowdisplayskip}{2pt} \setlength{\belowdisplayshortskip}{2pt}
\setlength{\abovedisplayskip}{2pt} \setlength{\abovedisplayshortskip}{2pt}
\begin{equation}\label{eq:loss_8}
	\begin{aligned}
		L_i = & L_i^{cls*} + L_i^{reg*}                                                                                                                            \\
		=     & \frac{1}{{\sigma^{cls}_i}^2} L_{i}^{cls} + \frac{1}{{\sigma^{reg}_i}^2} L_{i}^{reg} + \lambda_1 \log\sigma^{cls}_i + \lambda_2 \log\sigma^{reg}_i,
	\end{aligned}
\end{equation}
Note that directly predicting ${\sigma_i^{\cdot}}^2$ brings implementation difficulties since ${\sigma_i^{\cdot}}^2$ is expected to be positive and putting ${\sigma_i^{\cdot}}^2$ in the denominator position has the potential danger of division by zeros. Following \cite{kendall2018multi}, we instead predict ${m_i}^{\cdot}: = log (\sigma_i^{\cdot}) $, which makes the optimization more numerically stable and allows for unconstrained prediction output. Eventually, the overall loss function becomes:
\setlength{\belowdisplayskip}{2pt} \setlength{\belowdisplayshortskip}{2pt}
\setlength{\abovedisplayskip}{2pt} \setlength{\abovedisplayshortskip}{2pt}
\begin{equation}\label{eq:loss_9}
	\begin{aligned}
		L_{i} = & exp (-2 * m^{cls}_i) L_i^{cls} + \lambda_1 m^{cls}_i   \\
		+       & exp (-2 * m^{reg}_i) L_i^{reg} + \lambda_2 m^{reg}_i ,
	\end{aligned}
\end{equation}

\textbf{Theoretic analysis.} There exist two opposite sample weighting strategies for object detector training. On the one hand, some prefer ``hard'' samples, which can effectively accelerate the training procedure via a more significant magnitude of loss and gradient. On the other hand, some believe that ``easy'' examples need more attention when ranking is more important for evaluation metric and the class imbalance problem is superficial. However, it is usually not realistic to manually judge how hard or noisy a training sample is. Therefore, involving {\em{sample level}} variance as in Eq.~\ref{eq:loss_5} introduces more flexibility, as it allows adapting the sample weights automatically based on the effectiveness of each sample feature.

Taking derivatives of Eq.~\ref{eq:loss_5} with respect to the variance $\sigma^{{reg}}_{i}$, equating to zero and solving (assuming $\lambda_2=1$) , the optimal variance value satisfies ${{\sigma_i^{{{reg,*}}}}^2}=L_i^{{reg}}$ . Plugging this value back into Eq.~\ref{eq:loss_5} and ignoring constants, the overall regression objective reduces to $\log L_i^{{reg}}$. This function is a concave non-decreasing function that heavily favors $L_{i}^{reg}=||gt_i -a^*_i||_{2}^2\rightarrow 0$, while it applies only soft penalization for {\em large} $L_i^{{reg}} $ values. This makes the algorithm robust to outliers and noisy samples having large gradients that potentially degrade the performance. This also prevents the algorithm focusing too much on hard samples where $L_i^{reg}$ is drastically large. In this way, the regression function Eq.~\ref{eq:loss_5} favors a selection of samples having large IoUs as this encourages a faster speed that drives the loss towards minus infinity. This, in turn, creates an incentive for the feature learning procedure to {\em{weigh more}} on these samples, while samples having relatively smaller IoUs still maintain a modest gradient during the training.

Note that we have different weights $ (\exp (-2 * m^{cls}_i) $ and $ (\exp (-2 * m^{reg}_i) $ tailored for each sample. This is critical for our algorithm as it allows to adjust the multi-task balance weight at a sample level. In the next section, we describe how the loss function effectively drives the network to learn useful sample weights via our network design.

\subsection{Unified Sample Weighting Network Design}
Figure~\ref{fig:framework} shows the framework of our Sample Weighting Network (SWN). As we can see, the SWN is a sub-network of the detector supervised by detection objective, which takes some input features to predict weights for each sample. Our network is very simple, which consists of two levels of Multiple Layer Perception (MLP) networks as shown in Figure~\ref{fig:framework} (c). Instead of directly using the sample's visual feature, which actually misses the information from the corresponding ground truth, we design four discriminative features from the detector itself. It leverages the interaction between the estimation and the ground truth i.e., the IoU and classification score, because both classification and regression losses inherently reflect the prediction uncertainty to some extent.

More specifically, it adopts the following four features: the classification loss $L^{cls}_{i}$, the regression loss $L^{reg}_{i}$, $IoU_{i}$ and $Prob_{i}$, respectively, as an input. For negative samples, the $IoU_{i}$ and $Prob_{i}$ are set to 0. Next, we introduce four functions $F$, $G$, $H$ and $K$ to transform the inputs into dense features for a more comprehensive representation. These functions are all implemented by the MLP neural networks, which are able to map each one dimension value into a higher dimensional feature. We encapsulate those features into a sample-level feature $d_{i}$:
\setlength{\belowdisplayskip}{2pt} \setlength{\belowdisplayshortskip}{2pt}
\setlength{\abovedisplayskip}{2pt} \setlength{\abovedisplayshortskip}{2pt}
\begin{equation}\label{eq:loss_10}
	d_{i} = concat (F (L^{cls}_{i}) ;G (L^{reg}_{i}) ; H (IoU_{i}) ; K (Prob_{i})),
\end{equation}

In the upcoming step, the adaptive sample weight $m^{cls}_{i}$ for classification loss and $m^{reg}_{i}$ for regression loss are learned from the sample feature $d_i$, as follows:
\setlength{\belowdisplayskip}{2pt} \setlength{\belowdisplayshortskip}{2pt}
\setlength{\abovedisplayskip}{2pt} \setlength{\abovedisplayshortskip}{2pt}
\begin{equation}\label{eq:loss_11}
	m^{cls}_{i}= W_{cls} (d_{i}) ~~{\rm{and}}~~ m^{reg}_{i}= W_{reg} (d_{i}),
\end{equation}
where $W_{cls}$ and $W_{reg}$ represent two separate MLP networks for classification and regression weight prediction.

As shown in Figure~\ref{fig:framework}, our SWN has no assumption on the basic object detectors, which means it can work with most region-based object detectors, including Faster R-CNN, RetinaNet, and Mask R-CNN. To demonstrate the generalization of our method, we make minimal modifications to the original frameworks. Faster R-CNN consists of region proposal network (RPN) and Fast R-CNN network. We leave the RPN unchanged and plug the sample weighting network into the Fast R-CNN branch. For each sample, we firstly compute $L^{cls}_{i}$, $L^{reg}_{i}$, ${IoU}_{i}$, and $Prob_{i}$ as the inputs to the SWN. The predicted weights $\exp (-2*m^{cls}_{i})$ and $\exp (-2*m^{reg}_{i})$ are then inserted into Eq.~\ref{eq:loss_9} and the gradient is backpropagated to the base detection network and sample weighting network. For RetinaNet, we follow a similar process to generate the classification and regression weights for each sample. As Mask R-CNN has an additional mask branch, we include another branch into sample weighting network to generate adaptive weights for mask loss, where the classification, bounding box regression and mask prediction are jointly estimated. In order to match the additional mask weights, we also add the mask loss as an input to the sample weighting network.

In our experiments, we find that the predicted classification weights are not stable since the uncertainties among negative samples and positive samples are much more diverse than that of regression. Consequently, we average the classification weights of positive samples and negative samples separately in each batch, which can be viewed as a smooth version of weight prediction for classification loss.
\section{Experiments}
\begin{table*}[htb]
	\centering
	\caption{\small Results of different detectors on COCO \emph{test-dev}.}
	\vspace{-0.1in}
	\label{tab:results}
	\begin{tabular}{l|l| c | c c c c c c c}
		\hline
		Method                        & Backbone    & AP                              & $\text{AP}_{50}$ & $\text{AP}_{75}$ & $\text{AP}_{S}$ & $\text{AP}_{M}$ & $\text{AP}_{L}$      \\
		\hline
		\emph{Two-stage detectors}    &             &                                 &                  &                  &                 &                 &                 &  & \\
		Faster R-CNN                  & ResNet-50   & 36.7                            & 58.8             & 39.6             & 21.6            & 39.8            & 44.9                 \\
		Faster R-CNN                  & ResNeXt-101 & 40.3                            & 62.7             & 44.0             & 24.4            & 43.7            & 49.8                 \\
		Mask R-CNN                    & ResNet-50   & 37.5                            & 59.4             & 40.7             & 22.1            & 40.6            & 46.2                 \\
		Mask R-CNN                    & ResNeXt-101 & 41.4                            & 63.4             & 45.2             & 24.5            & 44.9            & 51.8                 \\
		\hline
		Faster R-CNN w/ SWN           & ResNet-50   & $\textbf{ 38.5}_{\uparrow 1.8}$ & 58.7             & 42.1             & 22.0            & 41.3            & 48.2                 \\
		Faster R-CNN w/ SWN           & ResNeXt-101 & $\textbf{41.4}_{\uparrow 1.1}$  & 61.9             & 45.3             & 24.1            & 44.7            & 52.0                 \\
		Mask R-CNN w/ SWN             & ResNet-50   & $\textbf{39.0}_{\uparrow 1.5}$  & 58.9             & 42.7             & 21.9            & 42.1            & 49.2                 \\
		Mask R-CNN w/ SWN             & ResNeXt-101 & $\textbf{42.5}_{\uparrow 1.1}$  & 64.1             & 46.6             & 24.8            & 46.0            & 53.5                 \\
		\hline
		\hline
		\emph{Single-stage detectors} &             &                                 &                  &                  &                 &                 &                 &  & \\
		RetinaNet                     & ResNet-50   & 35.9                            & 56.0             & 38.3             & 19.8            & 38.9            & 45.0                 \\
		RetinaNet                     & ResNeXt-101 & 39.0                            & 59.7             & 41.9             & 22.3            & 42.5            & 48.9                 \\
		\hline
		RetinaNet w/ SWN              & ResNet-50   & $\textbf{37.2}_{\uparrow 1.3}$  & 55.8             & 39.8             & 20.6            & 40.1            & 46.2                 \\
		RetinaNet w/ SWN              & ResNeXt-101 & $\textbf{40.8}_{\uparrow 1.8}$  & 60.1             & 43.8             & 23.2            & 44.0            & 51.1                 \\
		\hline
	\end{tabular}
	\vspace{-0.2in}
\end{table*}
We conducted thorough experiments on the challenging MS COCO \cite{lin2014microsoft} and Pascal VOC \cite{everingham2010pascal} datasets and evaluated our method with both one-stage and two-stage detectors.

\subsection{Datasets and Evaluation Metrics}
MS COCO \cite{lin2014microsoft} contains 80 common object categories in everyday scenes. Following the common practice, we used the \textit{train2017} split for training. It has 115k images and 860k annotated objects. We tested our approach as well as other compared methods on COCO \textit{test-dev} subset. Since the labels of \textit{test-dev} are not publicly available, we submitted all results to the evaluation server for evaluation. Yet all ablation experiments are evaluated on the \textit{val2017} subset which contains 5k images. Pascal VOC \cite{everingham2010pascal} covers 20 common categories in everyday life. We merged the \textit{VOC07 trainval} and \textit{VOC12 trainval} split for training and evaluated on \textit{VOC07 test} split.
Our evaluation metric is the standard COCO-style mean Average Precision (mAP) under different IoU thresholds, ranging from 0.5 to 0.95 with an interval of 0.05. It reflects detection performance under various criteria and favors high precisely localized detection results.

\subsection{Implementation Details}
We implemented our methods based on the publicly available mmdetection toolbox\cite{chen2019mmdetection}. In our experiments, all models were trained end-to-end with 4 Tesla P40 GPUs (each GPU holds 4 images) for 12 epochs, which is commonly referred as 1x training schedule. The base detection networks excluding the SWN is trained with stochastic gradient descent (SGD). The initial learning rate was set to 0.02 and decreased by 0.1 after epoch 8 and 11. For the sample weighting network, we adopted Adam \cite{kingma2014adam} with 0.001 learning rate and followed the same learning rate decay schedule as base detection network. The weight decay of 0.0001 was used for both optimizers. Other hyperparameters closely follow the settings in mmdetection unless otherwise specified. We initialized the weights of FC layers in the SWN with Gaussian distribution. The standard deviation and mean were set to 0.0001 and 0, and thus the predicted weights are nearly uniform across samples at the beginning of training. We also enforced the predicted weights to fall into the range of $[-2, 2]$ by clipping the values out of bounds, which stabilizes the training in practice. Faster R-CNN, Mask R-CNN and RetinaNet are chosen as the representative two-stage and one-stage detectors. Two classical networks, ResNet-50 and ResNext-101-32x4d are adopted as backbones and FPN is used by default. \textit{Please note that our method is fairly general and thus not limited to the aforementioned detectors and backbones.} In fact, it is applicable to any two-stage and one-stage detectors and is transparent to the choice of backbone networks.

\subsection{Results}
As discussed, our sample weighting network (SWN) can be applied to any region-based object detector. To verify the effectiveness of our method for performance boosting, we evaluated it thoroughly on Faster R-CNN, Mask R-CNN and RetinaNet (one of the latest one-stage detectors performing better than SSD) with two backbones ResNet-50 and ResNeXt-101-32x4d. Table~\ref{tab:results} shows the results on COCO \emph{test-dev} in terms of Average Precision (AP). Thanks to the proposed SWN, all detectors have achieved consistent performance gains up to 1.8\%. Especially, the boost to RetinaNet is very impressive because it already has a strong sample weighting strategy. All improvements indicate that our SWN is complementary to the detectors' internal sample weighting strategies. In addition, from column $AP_{S}$, $AP_{M}$ and $AP_{L}$ (AP results for small, medium and large objects respectively) , we notice that our weighting strategy works better for ``large'' objects. Furthermore, we can infer from the results that the AP boosts are larger at higher IoU.

It is worth mentioning that SWN only affects the detector training with minimal extra cost. As an example, adding SWN to ``Faster R-CNN + ResNet-50'' detector only increased the training time from 1.009s to 1.024s per iteration and parameters from 418.1M to 418.4M. More importantly, since the inference is exactly the same, our approach does not add any additional cost to the test, which makes our sampling strategy more practical.

We also conducted similar evaluations on the PASCAL VOC 2007 dataset. The experimental reports are summarized in Table~\ref{tab:voc-results}. In terms of AP, our approach further demonstrates its effectiveness on performance improvements. According to the gains on both popular benchmark datasets, we can believe our SWN can consistently boost the performance of region-based object detectors.

\begin{table}[t]
	\centering
	\caption{\small Results of different detectors on VOC2007 \emph{test}.}
	\addtolength{\tabcolsep}{-5pt}
	\vspace{-0.1in}
	\label{tab:voc-results}
	\begin{tabular}{l|l|c}
		\hline
		Method                        & Backbone    & AP                             \\
		\hline
		\emph{Two-stage detectors}    &             &                                \\
		Faster R-CNN                  & ResNet-50   & 51.0                           \\
		Faster R-CNN                  & ResNeXt-101 & 54.2                           \\
		Faster R-CNN w/ SWN           & ResNet-50   & $\textbf{52.5}_{\uparrow 1.5}$ \\
		Faster R-CNN w/ SWN           & ResNeXt-101 & $\textbf{56.0}_{\uparrow 1.8}$ \\ \hline\hline
		\emph{Single-stage detectors} &             &                                \\
		RetinaNet                     & ResNet-50   & 52.0                           \\
		RetinaNet                     & ResNeXt-101 & 55.3                           \\
		RetinaNet w/ SWN              & ResNet-50   & $\textbf{53.4}_{\uparrow 1.4}$ \\
		RetinaNet w/ SWN              & ResNeXt-101 & $\textbf{56.8}_{\uparrow 1.5}$ \\
		\hline
	\end{tabular}
	\vspace{-0.2in}
\end{table}

\begin{figure*}[htb]
	\centering {\includegraphics[width=1.0\textwidth]{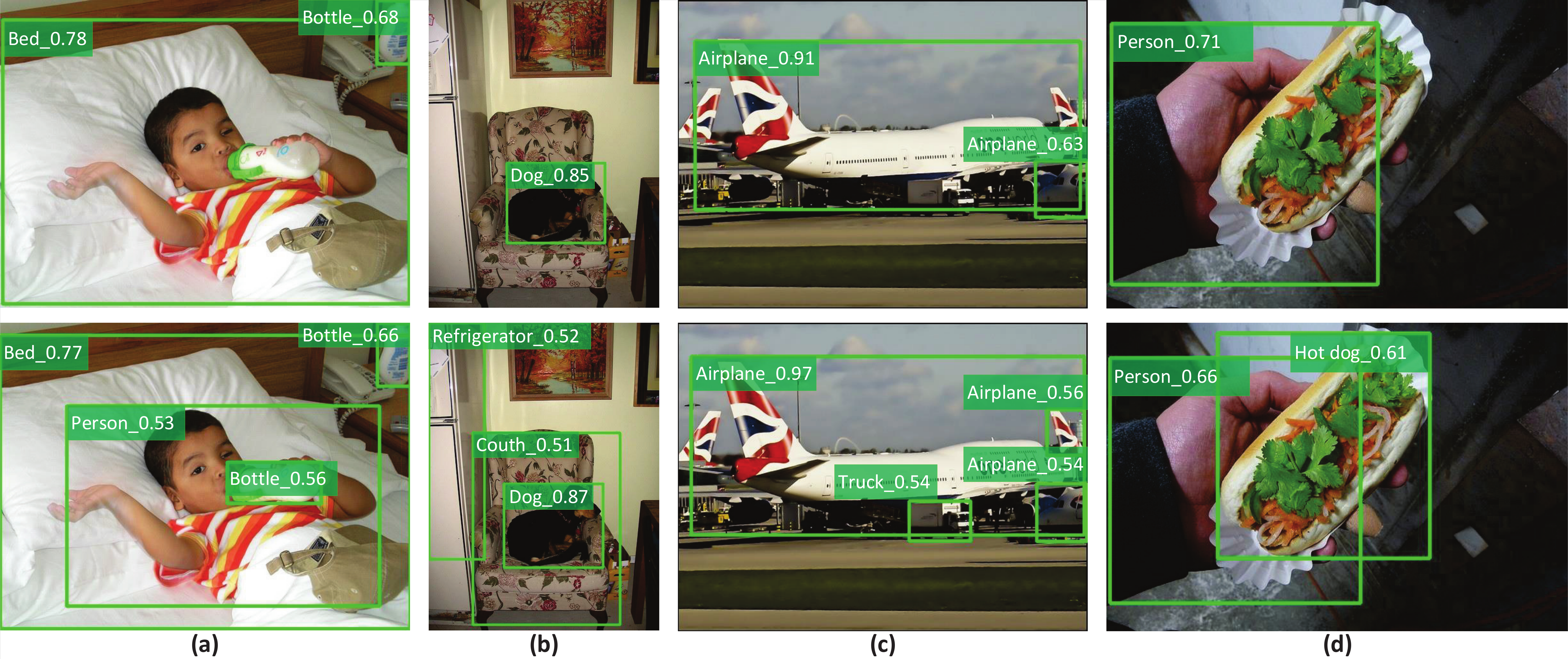}}
	\vspace{-0.1in}
	\caption{\small Examples of detection results of RetinaNet (first row) and RetinaNet w/SWN (second row). RetinaNet missed detecting some ``easy'' objects such as a child, a coach, a hot dog, and etc., which have been successfully detected by the boosted RetinaNet with SWN.}
	\label{fig:tech_1}
\end{figure*}

Figure~\ref{fig:tech_1} demonstrates some qualitative performance comparisons between RetinaNet and RetinaNet+SWN on COCO dataset. Following a common threshold of 0.5 used for visualizing detected objects, we only illustrate a detection when its score is higher than the threshold. As we can see, some so-called ``easy'' objects such as a child , a coach, a hot dog and so on, which are missed by RetinaNet, have been successfully detected by the boosted RetinaNet with SWN. We conjecture that original RetinaNet may concentrate too much on ``hard'' samples. As a result, the ``easy'' samples get less attention and make less contributions to the model training. The scores for these ``easy'' examples have been depressed, which results in the missing detections. The purpose of Figure~\ref{fig:tech_1} is not to show the ``bad" of RetinaNet in score calibration, because the ``easy" ones can be detected anyway when decreasing the threshold. Figure~\ref{fig:tech_1} actually illustrates that unlike RetinaNet, SWN doesn't weigh less on ``easy'' examples.

There is another line of research, which aims to improve bounding box regression. In other words, they attempt to optimize the regression loss by learning with IoU as the supervision or its combination with NMS. Based on the Faster R-CNN + ResNet-50 + FPN framework, we make a comparison on COCO \emph{val2017} as shown in Table~\ref{tab:reweighting}. The performance comparison shows that both our SWN and its extension SWN+Soft-NMS outperform the IoU-Net and IoU-Net+NMS. It further confirms the advantages of learning sample weights for both classification and regression.

\begin{table} [tb]
	\centering
	\caption{\small Performance comparisons with IoU-based approaches.}
	\label{tab:reweighting}
	\vspace{-0.1in}
	\addtolength{\tabcolsep}{-5pt}
	\begin{tabular}{c| c c c c c c}
		\hline
		                                        & AP            & AP$_{50}$ & AP$_{75}$ & AP$_\text{S}$ & AP$_\text{M}$ & AP$_\text{L}$ \\
		\hline
		Baseline                                & 36.4          & 58.4      & 39.1      & 21.6          & 40.1          & 46.6          \\
		IoU-Net \cite{jiang2018acquisition}     & 37.0          & 58.3      & -         & -             & -             & -             \\
		IoU-Net+NMS \cite{jiang2018acquisition} & 37.6          & 56.2      & -         & -             & -             & -             \\
		SWN                                     & 38.2          & 58.1      & 41.6      & 21.3          & 41.7          & 50.2          \\
		SWN + Soft-NMS                          & \textbf{39.2} & 58.6      & 43.3      & 22.3          & 42.6          & 51.1          \\
		\hline
	\end{tabular}
	\vspace{-0.1in}
\end{table}

\begin{table}[tb]
	\centering
	\caption{\small Effectiveness of each component.}
	\vspace{-0.1in}
	\label{tab:components}
	\addtolength{\tabcolsep}{-2pt}
	\begin{tabular}{cc|cccccc}
		\hline
		CLS        & REG        & AP                            & AP$_{50}$ & AP$_{75}$ & AP$_\text{S}$ & AP$_\text{M}$ & AP$_\text{L}$ \\
		\hline
		           &            & 36.4                          & 58.4      & 39.1      & 21.6          & 40.1          & 46.6          \\
		\checkmark &            & $36.7_{\uparrow0.3}$          & 58.7      & 39.5      & 21.2          & 40.2          & 47.9          \\
		           & \checkmark & $37.0_{\uparrow0.6}$          & 56.6      & 40.1      & 21.2          & 40.4          & 47.9          \\
		\checkmark & \checkmark & $\textbf{38.2}_{\uparrow1.8}$ & 58.1      & 41.6      & 21.3          & 41.7          & 50.2          \\
		\hline
	\end{tabular}
	\vspace{-0.2in}
\end{table}

\subsection{Ablation Study and Analysis}
For a better understanding to our SWN, we further conducted a series of ablation studies on COCO \emph{val2017} using Faster R-CNN + ResNet-50 as our baseline.

The first group of experiments we did is to verify how well our approach works for each individual task, i.e., object classification (CLS) and regression (REG). Table~\ref{tab:components} shows the detailed results. If a component is selected, it means our weighting strategy has been applied to it. The results clearly demonstrate that when the sample weighting is applied to only one task, the performance boost is trivial. Nonetheless, jointly applying it to both tasks can achieve a significant performance improvement of 1.8\%. This observation is consistent with the goal of our SWN design.

There are two regularization hyperparameters (i.e., $\lambda_{1}$ and $\lambda_{2}$) in our loss function. In this set of experiments, we assigned various values to these parameters to check how sensitive our approach is to different regularization magnitudes. In our implementation, two parameters always share the same value. Table~\ref{tab:loss_weight} illustrates the comparisons. It shows the results are relatively stable when $\lambda$ lies in the range of 0.3 and 0.7, and achieves best performance at 0.5.

\begin{table} [tb]
	\centering
	\vspace{-0.1in}
	\caption{\small Performance comparisons by varying $\lambda$.}
	\vspace{-0.1in}
	\label{tab:loss_weight}
	\addtolength{\tabcolsep}{3pt}
	\begin{tabular}{c| c c c c c }
		\hline
		$\lambda$ & 0.1  & 0.3  & 0.5           & 0.7  & 1.0  \\
		\hline
		AP        & 29.3 & 37.4 & \textbf{38.2} & 37.9 & 37.2 \\
		\hline
	\end{tabular}
	\vspace{-0.1in}
\end{table}

\begin{figure}[t]
	\vspace{-0.1in}
	\centering {\includegraphics[width=0.45\textwidth]{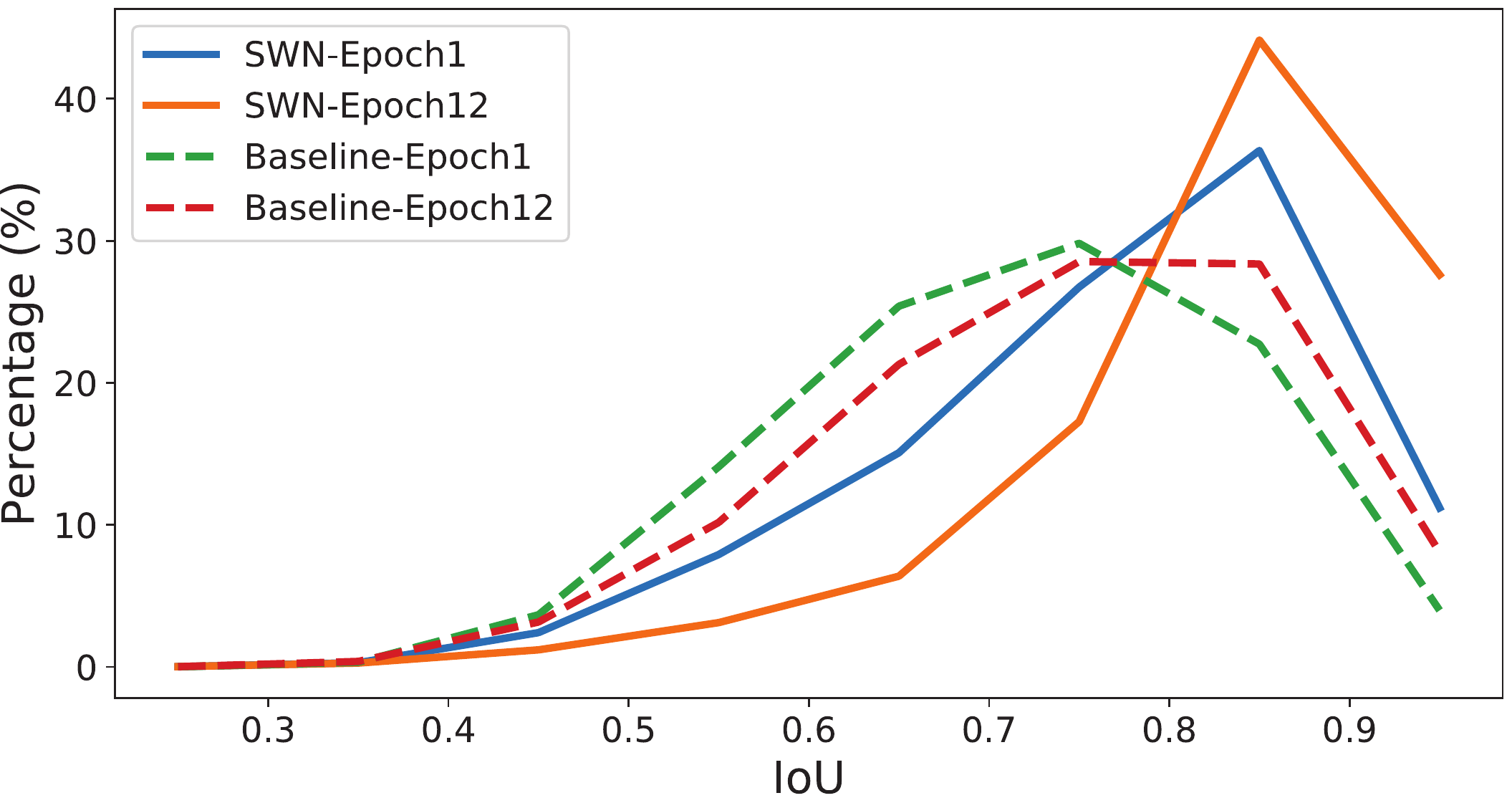}}
	\caption{\small Classification loss distribution of positive samples with different IoUs. Higher IoUs mean easier samples. Y-axis denotes the percentage of weighted loss. For example, percentage=20\% at IoU=0.85 with SWN-Epoch12 means the the losses of samples whose IoUs fall between 0.8 and 0.9 take up 42\% of total loss.}
	\label{fig:cls_loss_dist}
	\vspace{-0.2in}
\end{figure}

To understand the learning process, we draw the distribution of classification loss over samples at different IoUs as shown in Figure~\ref{fig:cls_loss_dist}. We picked up the data from two training epochs to derive the distributions for both Baseline and SWN. The x-axis represents samples at a certain IoU with ground truth. Samples with higher IoUs shall have less uncertainties and thus higher weights to be considered by the loss optimization. There are two observations from the distributions. First, during the optimization process, the classification loss will draw more attentions to ``easy'' samples (i.e., the ones with high IoU values). Second, our approach generally put more weights to samples with high IoU values when computing the loss. All observations are consistent with our previous analysis of SWN.

\section{Conclusion}

We have demonstrated that the problem of sample weighting for region-based object detection is both data-dependent and task-dependent. The importance of a sample to detection optimization is also determined by its uncertainties shown in two correlated classification and regression losses. We derive a general principled loss function which can automatically learn sample-wise task weights from the training data. It is implemented with a simple yet effective neural network, which can be easily plugged into most region-based detectors without additional cost to inference. The proposed approach has been thoroughly tested on different datasets, and consistent performance gains up to 1.8\% have been observed. Some qualitative results clearly illustrate that our approach can detect some ``easy'' objects which are missed by other detectors. In future work, we will work on a complete explanation of this phenomenon. In addition, we can continue to improve our approach such that it can deal with ``hard'' and ``easy'' samples more smartly at different optimization phrases.

\appendix
\section{Appendix}

\renewcommand\thefigure{\thesection.\arabic{figure}}
\renewcommand\thetable{\thesection.\arabic{table}}
\setcounter{figure}{0}
\setcounter{table}{0}

\setcounter{table}{0}
\renewcommand{\thetable}{A.\arabic{table}}

This supplementary material contains 1) the derivation and the approximation of classification loss in Eq. 7 of the main paper; 2) the details of the average of classification weights; 3) the regression loss distribution over samples at different IoUs; 4) the convergence progress of the sample weights and losses during training; 5) the sensitivity of SWN with respect to different initializations. We adopt Faster R-CNN + ResNet-50 + FPN as our baseline in the experiments of this material.

\subsection{Technical Details}
\subsubsection{Derivation of Classification Loss in Eq. 7}
In Section 3.3 of the main paper, we approximate $L_i^{cls*}$ with $\frac{1}{{\sigma^{cls}_i}^2} L_{i}^{cls} + \lambda_1 \log\sigma^{cls}_i$. Here we present the detailed derivation. By taking the log on Eq. 6 of the main paper and multiplying with -1, we obtain the $L_i^{cls*}$ as follows:
\begin{equation}\label{supp:eq_1}\scriptsize
	\begin{aligned}
		  & L_i^{cls*}= -\log\bigg[ softmax( y_i, \frac{1}{{  {\sigma_i^{cls}  }^ 2}}p(a^*_i))       \bigg]                                                                                                                                                                                                                                                                                                                                                                                            \\
		= & -\log \bigg[ \frac{   \exp( \frac{1}{{  {\sigma_i^{cls}  }^ 2}} p(a^*_i)_{y_i} ) } {  \sum_{c^\prime}  \exp(     {\frac{1}{{  {\sigma_i^{cls}  }^ 2}}} {p(a^*_i)}_{c^{\prime}} ) }  \bigg]                                                                                                                                                                                                                                                                                                 \\
		= & -\log \bigg[ \frac{   (\exp( p(a^*_i)_{y_i} ) ) ^ { { \frac{1}{{  {\sigma_i^{cls}  }^ 2}} } } }{      ( \sum_{c^\prime}      (\exp(p(a^*_i)_{c^\prime} ) ) )^ { { \frac{1}{{  {\sigma_i^{cls}  }^ 2}} } }           }  \cdot   \frac{    ( \sum_{c^\prime}      (\exp(  p(a^*_i)_{c^\prime} ) ) )^ { {    \frac{1}{{  {\sigma_i^{cls}  }^ 2}}  } }   }{             {  \sum_{c^\prime}  \exp(  {\frac{1}{{  {\sigma_i^{cls}  }^ 2}}} {p(a^*_i)}_{c^{\prime}}  ) }                  }\bigg] \\
		= & -\frac{1}{{  {\sigma_i^{cls}  }^ 2}}        \log\bigg[ softmax( y_i, p(a^*_i))\bigg]      - \log\bigg[      \frac{    ( \sum_{c^\prime}      (\exp(p(a^*_i)_{c^\prime} ) ) )^ { {    \frac{1}{{  {\sigma_i^{cls}  }^ 2}}  } }   }{             {  \sum_{c^\prime}  \exp(  {\frac{1}{{  {\sigma_i^{cls}  }^ 2}}} {p(a^*_i)}_{c^{\prime}}  ) }                  }         \bigg].
	\end{aligned}
\end{equation}
Similar to \cite{kendall2017uncertainties}, we approximate the last term in Eq. \ref{supp:eq_1} with:
\begin{equation}\label{supp:eq_2}\small
	\frac{1}{{  {\sigma_i^{cls}  }}} \sum_{c^\prime}  exp(  {\frac{1}{{  {\sigma_i^{cls}  }^ 2}}} {p(a^*_i)}_{c^{\prime}}  )  \approx    ( \sum_{c^\prime}      (exp(  p(a^*_i)_{c^\prime} ) ) )^ { {    \frac{1}{{  {\sigma_i^{cls}  }^ 2}}  } }.
\end{equation}

The approximation becomes an equality when $\sigma_i^{cls}\rightarrow1$. By inserting the definition of $L_i^{cls}$ back into Eq. \ref{supp:eq_1} and under the approximation of Eq. \ref{supp:eq_2}, we obtain:
\begin{equation}\label{supp:eq_3}\small
	L_i^{cls*} = \frac{1}{{\sigma^{cls}_i}^2} L_{i}^{cls} + \log\sigma^{cls}_i.
\end{equation}
The second term in Eq. \ref{supp:eq_3} can be regarded as a regularization term that prevents the estimated weights from collapsing to zeros. The scales between $L_i^{cls}$ and  $\log\sigma^{cls}_i$ determine the optimal value of $\log\sigma^{cls}_i$ and further the classification weight $\frac{1}{{\sigma^{cls}_i}^2} $. To constrain the weight in a reasonable range, we introduce a hyper-parameter $\lambda_1$ in order to control the influence of  $\log\sigma^{cls}_i$:
\begin{equation}\label{supp:eq_4}\small
	L_i^{cls*} = \frac{1}{{\sigma^{cls}_i}^2} L_{i}^{cls} + \lambda_1 \log\sigma^{cls}_i.
\end{equation}
Please note that introducing $\lambda_1$ and associating it with $\log\sigma^{cls}_i$ is essentially equivalent to scaling the loss weight of $L_{i}^{cls}$ in Eq. \ref{supp:eq_3} by $\frac{1}{\lambda_1}$, which is a fairly common adopted technique when training object detectors.

\subsubsection{The Details of Average of Classification Weights}
For each batch, suppose we have a set of classification weights for positive samples $\mathcal{S}^{cls}_{{P}} = \{s_i^{cls}| a_i \in {\mathcal{A}^P} \}$ and negative samples $\mathcal{S}_{{N}}^{cls} = \{s_i^{cls}| a_i \in {\mathcal{A}^N} \}$, respectively. For positive samples, we empirically set $s_i^{cls*} = \frac{1}{ | \mathcal{A}^P|}  \sum_{ s_i^{cls} \in \mathcal{S}^{cls}_{{P}} }  s_i^{cls}$, where $| \mathcal{A}^P|$ defines the number of positive samples in each batch. Likewise, we have  $s_i^{cls*} = \frac{1}{ | \mathcal{A}^N|}  \sum_{ s_i^{cls} \in \mathcal{S}^{cls}_{{N}} }  s_i^{cls}$ for negative samples, where $| \mathcal{A}^N|$ represents the number of negative samples in each batch. This modified version of sample weights can be viewed as a smoothed version of weight prediction which is \emph{batch-wise} rather than \emph{sample-wise}.

\subsection{Experimental Results}

\subsubsection{Regression Loss Distribution}
\begin{figure}[h]
	\vspace{0.15in}
	\centering {\includegraphics[width=0.48\textwidth]{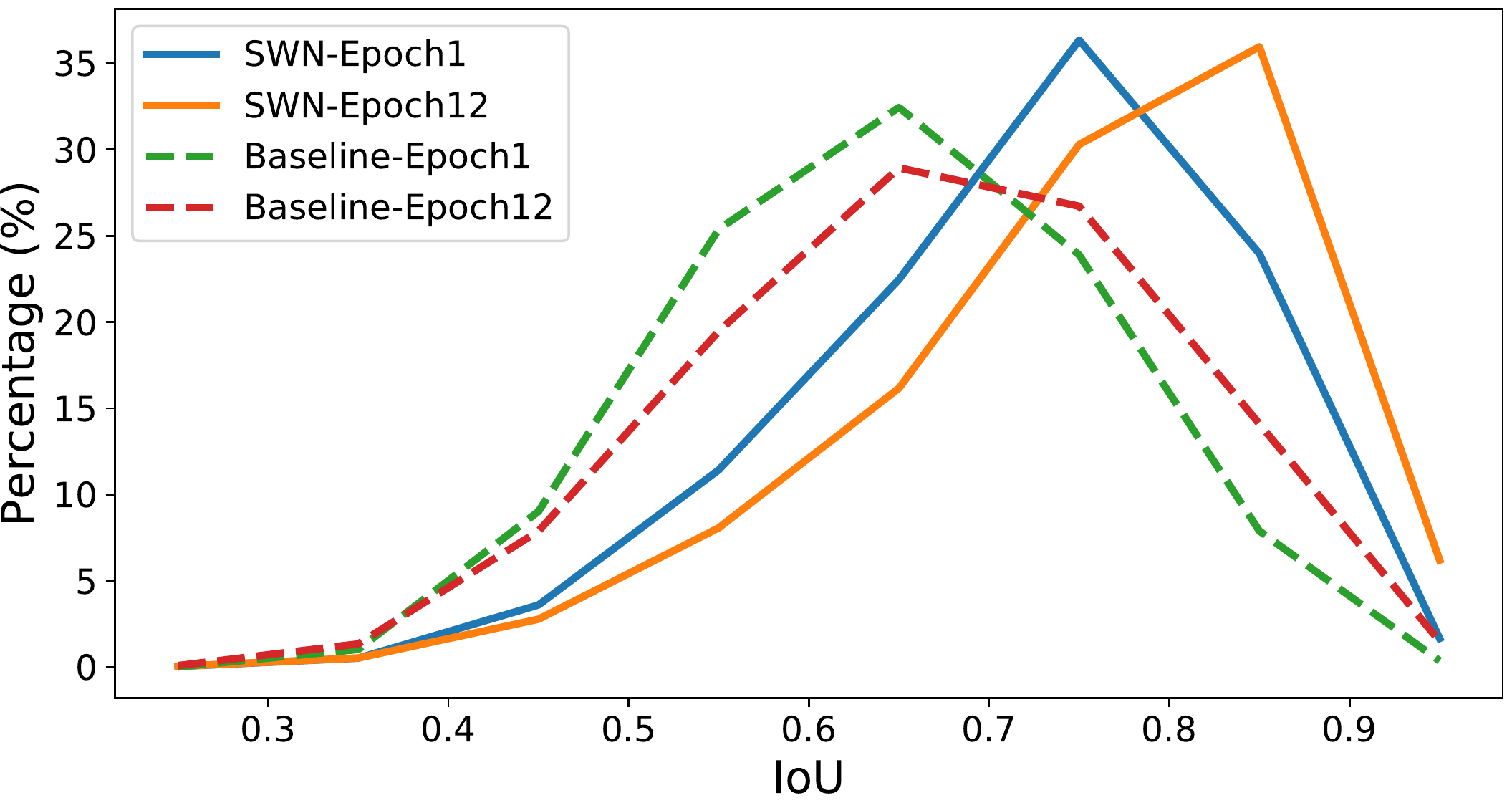}}
	\vspace{-0.2in}
	\caption{\small Regression loss distribution of positive samples with different IoUs.  The x-axis represents samples at a certain IoU with regard to its ground truth and y-axis denotes the percentage of regression loss. For example, percentage=35\% at IoU=0.85 with SWN-Epoch12 means that the loss of samples whose IoUs fall between 0.8 and 0.9 contributes 35\% of the total loss.}
	\vspace{-0.15in}
	\label{fig:reg_loss_dist}
\end{figure}

To further understand the learning process, we depict the distribution of regression losses at different IoUs as shown in Figure \ref{fig:reg_loss_dist}. We collect the data from epoch 1 and epoch 12 to derive the distributions for both baseline and our Sample Weighting Network (SWN). The x-axis represents samples at a certain IoU with regard to its ground truth and y-axis denotes the percentage of regression loss.
The samples with higher IoUs usually have less uncertainties and thus such samples weigh more on the loss of the optimization in training.
It can be observed that the regression loss will draw more attention to ``easy'' samples (i.e., the ones with high IoU values) during the optimization process. Additionally, our approach generally puts more weights on samples with high IoU values compared to baseline. All the observations nicely verify our analysis on SWN and are also consistent with that on classification loss distribution.

\subsubsection{Convergences of Sample Weights and Losses}
\begin{figure}[h]
	\centering {\includegraphics[width=0.48\textwidth]{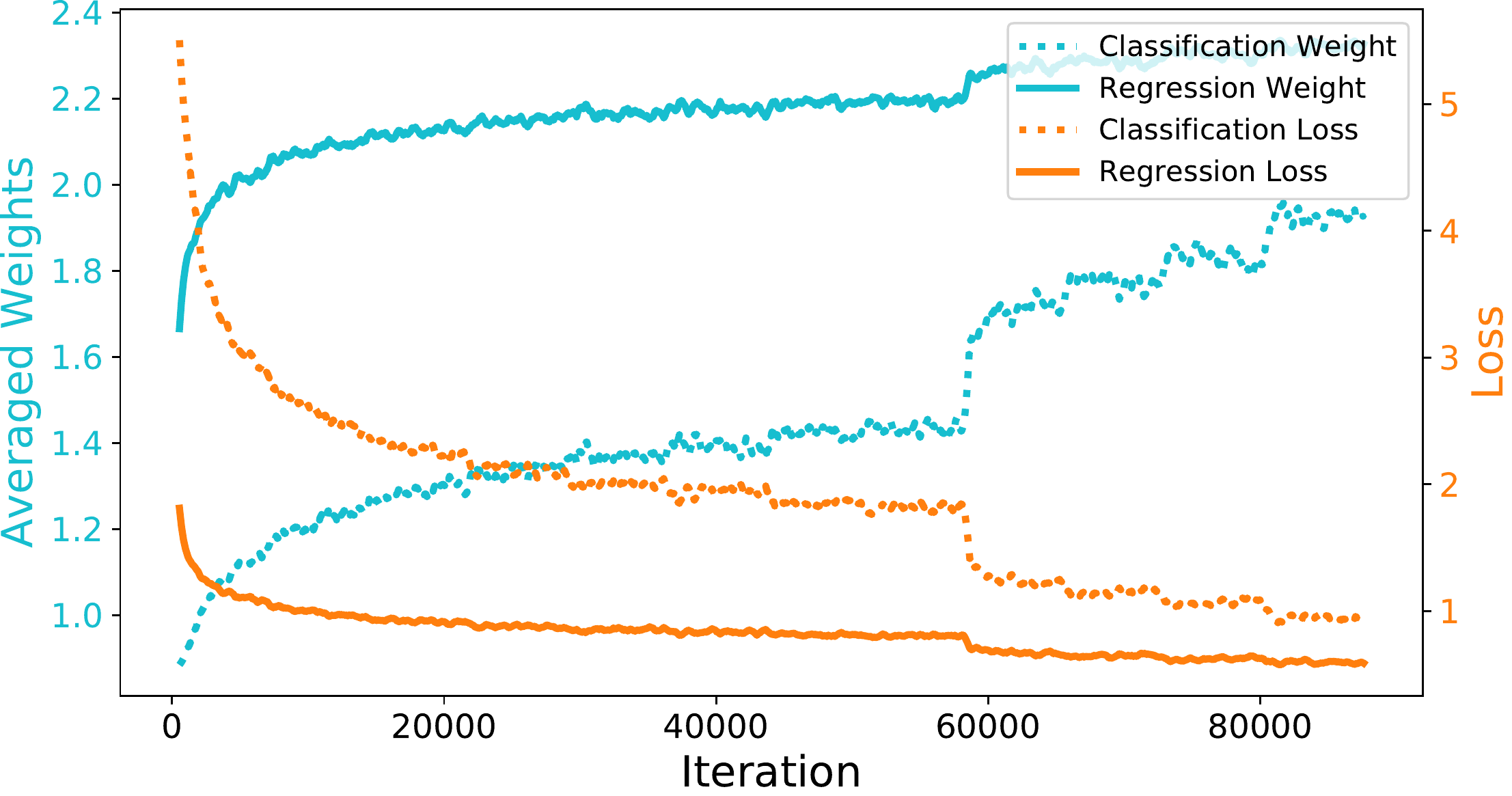}}
	\vspace{-0.2in}
	\caption{\small Classification losses, regression losses and sample weights during training. As the training proceeds, the sample weights increase while training losses decrease.}
	\vspace{-0.15in}
	\label{fig:training_iteration}
\end{figure}

Figure \ref{fig:training_iteration} shows convergences of classification losses, regression losses and sample weights during the training of Faster R-CNN + ResNet-50 + FPN with SWN.
The samples in the iterations at the late stage of training usually have less uncertainties and thus such samples weigh more on the loss in the optimization.
We can observe that as the training proceeds, the sample weights increase while the training losses decrease. This phenomenon again validates our analysis that the learning procedure \emph{weighs more} on easy (small training loss) samples.

\begin{figure}[t]
	\centering {\includegraphics[width=0.48\textwidth]{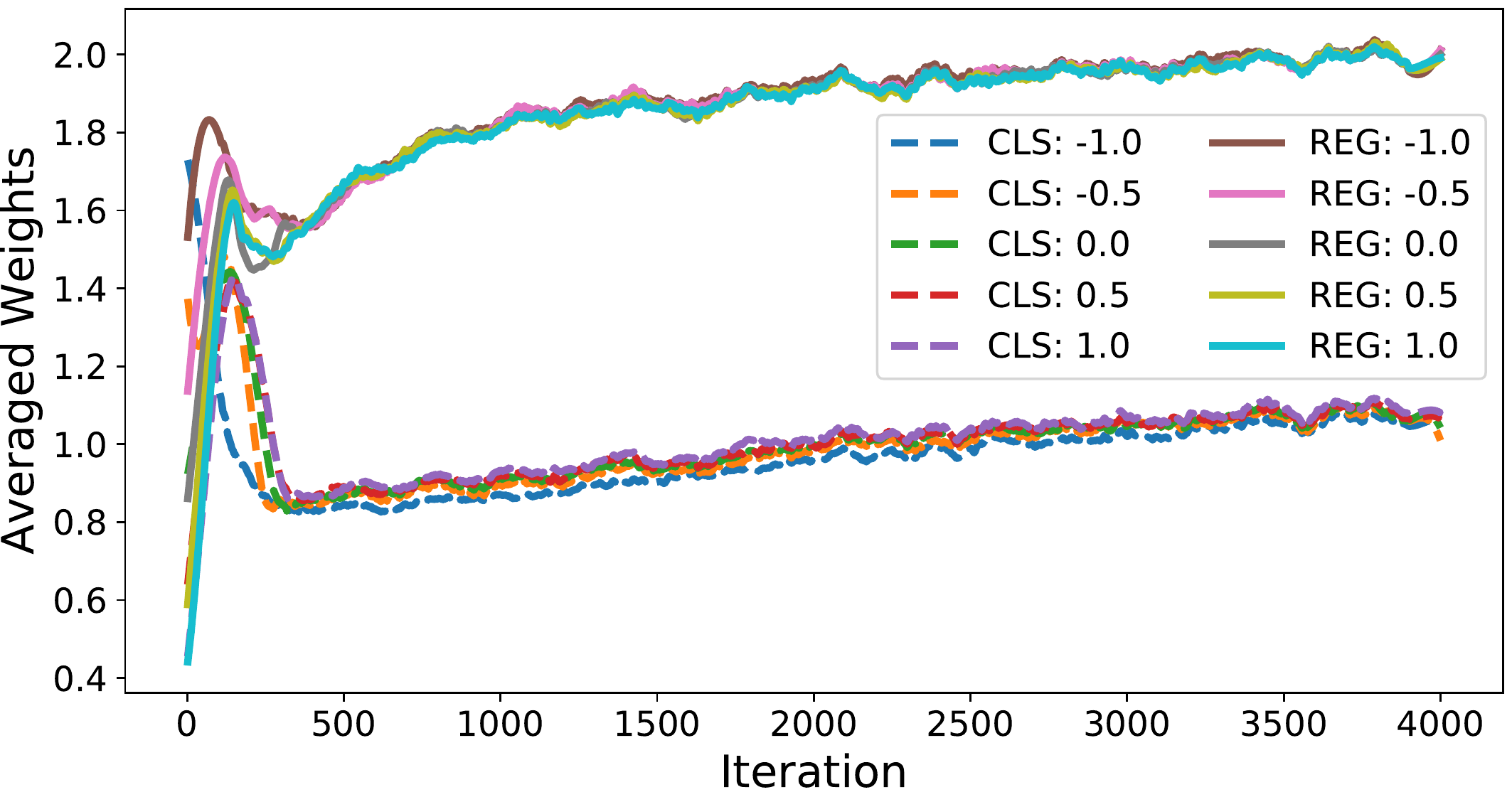}}
	\caption{\small Averaged classification weights and regression weights with different initializations. CLS represents averaged classification weights and REG represents averaged regression weights. The optimization processes lead to similar values of averaged weights after 500 iterations. Such results indicate that our model is robust to different initializations.}
	\label{supp:sensity}
\end{figure}

\subsubsection{Sensitivity of SWN w.r.t Different Initializations}
To investigate the sensitivity to different initializations, we set the bias of the last FC layer in SWN to different values and plot the averaged sample weights over iterations. Figure \ref{supp:sensity} shows that for a set of initial values ranging from -1.0 to 1.0 with an interval of 0.5, the optimization processes lead to similar values of averaged weights after 500 iterations.
The results indicate the robustness of our approach with respect to different initializations.

{\small
\bibliographystyle{ieee_fullname}
\bibliography{egbib}

\begin{thebibliography}{10}\itemsep=-1pt

\bibitem{cai2019exploring}
Qi Cai, Yingwei Pan, Chong-Wah Ngo, Xinmei Tian, Lingyu Duan, and Ting Yao.
\newblock Exploring object relation in mean teacher for cross-domain detection.
\newblock In {\em CVPR}, 2019.

\bibitem{cai2018cascade}
Zhaowei Cai and Nuno Vasconcelos.
\newblock Cascade r-cnn: Delving into high quality object detection.
\newblock In {\em CVPR}, 2018.

\bibitem{cao2019prime}
Yuhang Cao, Kai Chen, Chen~Change Loy, and Dahua Lin.
\newblock Prime sample attention in object detection.
\newblock {\em arXiv preprint arXiv:1904.04821}, 2019.

\bibitem{chen2019hybrid}
Kai Chen, Jiangmiao Pang, Jiaqi Wang, Yu Xiong, Xiaoxiao Li, Shuyang Sun,
  Wansen Feng, Ziwei Liu, Jianping Shi, Wanli Ouyang, et~al.
\newblock Hybrid task cascade for instance segmentation.
\newblock In {\em CVPR}, 2019.

\bibitem{chen2019mmdetection}
Kai Chen, Jiaqi Wang, Jiangmiao Pang, Yuhang Cao, Yu Xiong, Xiaoxiao Li,
  Shuyang Sun, Wansen Feng, Ziwei Liu, Jiarui Xu, et~al.
\newblock Mmdetection: Open mmlab detection toolbox and benchmark.
\newblock {\em arXiv preprint arXiv:1906.07155}, 2019.

\bibitem{chen2018domain}
Yuhua Chen, Wen Li, Christos Sakaridis, Dengxin Dai, and Luc Van~Gool.
\newblock Domain adaptive faster r-cnn for object detection in the wild.
\newblock In {\em CVPR}, 2018.

\bibitem{chen2019detnas}
Yukang Chen, Tong Yang, Xiangyu Zhang, Gaofeng Meng, Chunhong Pan, and Jian
  Sun.
\newblock Detnas: Neural architecture search on object detection.
\newblock In {\em NeurIPS}, 2019.

\bibitem{dai2016r}
Jifeng Dai, Yi Li, Kaiming He, and Jian Sun.
\newblock R-fcn: Object detection via region-based fully convolutional
  networks.
\newblock In {\em NeurIPS}, 2016.

\bibitem{dai2017deformable}
Jifeng Dai, Haozhi Qi, Yuwen Xiong, Yi Li, Guodong Zhang, Han Hu, and Yichen
  Wei.
\newblock Deformable convolutional networks.
\newblock In {\em ICCV}, 2017.

\bibitem{deng2019relation}
Jiajun Deng, Yingwei Pan, Ting Yao, Wengang Zhou, Houqiang Li, and Tao Mei.
\newblock Relation distillation networks for video object detection.
\newblock In {\em ICCV}, 2019.

\bibitem{everingham2010pascal}
Mark Everingham, Luc Van~Gool, Christopher~KI Williams, John Winn, and Andrew
  Zisserman.
\newblock The pascal visual object classes (voc) challenge.
\newblock {\em IJCV}, 2010.

\bibitem{fan2018learning}
Yang Fan, Fei Tian, Tao Qin, Xiang-Yang Li, and Tie-Yan Liu.
\newblock Learning to teach.
\newblock {\em ICLR}, 2018.

\bibitem{felzenszwalb2010DPM}
Pedro~F Felzenszwalb, Ross~B Girshick, and David McAllester.
\newblock Cascade object detection with deformable part models.
\newblock In {\em CVPR}, 2010.

\bibitem{freund1997decision}
Yoav Freund and Robert~E Schapire.
\newblock A decision-theoretic generalization of on-line learning and an
  application to boosting.
\newblock {\em JCSS}, 1997.

\bibitem{girshick2015fast}
Ross Girshick.
\newblock Fast r-cnn.
\newblock In {\em ICCV}, 2015.

\bibitem{girshick2014rich}
Ross Girshick, Jeff Donahue, Trevor Darrell, and Jitendra Malik.
\newblock Rich feature hierarchies for accurate object detection and semantic
  segmentation.
\newblock In {\em CVPR}, 2014.

\bibitem{he2017mask}
Kaiming He, Georgia Gkioxari, Piotr Doll{\'a}r, and Ross Girshick.
\newblock Mask r-cnn.
\newblock In {\em ICCV}, 2017.

\bibitem{he2016resnet}
Kaiming He, Xiangyu Zhang, Shaoqing Ren, and Jian Sun.
\newblock Deep residual learning for image recognition.
\newblock In {\em CVPR}, 2016.

\bibitem{he2019bounding}
Yihui He, Chenchen Zhu, Jianren Wang, Marios Savvides, and Xiangyu Zhang.
\newblock Bounding box regression with uncertainty for accurate object
  detection.
\newblock In {\em CVPR}, 2019.

\bibitem{hu2018relation}
Han Hu, Jiayuan Gu, Zheng Zhang, Jifeng Dai, and Yichen Wei.
\newblock Relation networks for object detection.
\newblock In {\em CVPR}, 2018.

\bibitem{jiang2018acquisition}
Borui Jiang, Ruixuan Luo, Jiayuan Mao, Tete Xiao, and Yuning Jiang.
\newblock Acquisition of localization confidence for accurate object detection.
\newblock In {\em ECCV}, 2018.

\bibitem{jiang2017mentornet}
Lu Jiang, Zhengyuan Zhou, Thomas Leung, Li-Jia Li, and Li Fei-Fei.
\newblock Mentornet: Learning data-driven curriculum for very deep neural
  networks on corrupted labels.
\newblock {\em ICML}, 2018.

\bibitem{kendall2017uncertainties}
A. Kendall and Y. Gal.
\newblock What uncertainties do we need in bayesian deep learning for computer
  vision?
\newblock In {\em NeurIPS}, 2017.

\bibitem{kendall2018multi}
Alex Kendall, Yarin Gal, and Roberto Cipolla.
\newblock Multi-task learning using uncertainty to weigh losses for scene
  geometry and semantics.
\newblock In {\em CVPR}, 2018.

\bibitem{khodabandeh2019robust}
Mehran Khodabandeh, Arash Vahdat, Mani Ranjbar, and William~G Macready.
\newblock A robust learning approach to domain adaptive object detection.
\newblock In {\em ICCV}, 2019.

\bibitem{kingma2014adam}
Diederik~P Kingma and Jimmy Ba.
\newblock Adam: A method for stochastic optimization.
\newblock {\em ICLR}, 2015.

\bibitem{kumar2010self}
M~Pawan Kumar, Benjamin Packer, and Daphne Koller.
\newblock Self-paced learning for latent variable models.
\newblock In {\em NeurIPS}, 2010.

\bibitem{lin2017feature}
Tsung-Yi Lin, Piotr Doll{\'a}r, Ross Girshick, Kaiming He, Bharath Hariharan,
  and Serge Belongie.
\newblock Feature pyramid networks for object detection.
\newblock In {\em CVPR}, 2017.

\bibitem{lin2017focal}
Tsung-Yi Lin, Priya Goyal, Ross Girshick, Kaiming He, and Piotr Doll{\'a}r.
\newblock Focal loss for dense object detection.
\newblock In {\em ICCV}, 2017.

\bibitem{lin2014microsoft}
Tsung-Yi Lin, Michael Maire, Serge Belongie, James Hays, Pietro Perona, Deva
  Ramanan, Piotr Doll{\'a}r, and C~Lawrence Zitnick.
\newblock Microsoft coco: Common objects in context.
\newblock In {\em ECCV}, 2014.

\bibitem{liu2016ssd}
Wei Liu, Dragomir Anguelov, Dumitru Erhan, Christian Szegedy, Scott Reed,
  Cheng-Yang Fu, and Alexander~C Berg.
\newblock Ssd: Single shot multibox detector.
\newblock In {\em ECCV}, 2016.

\bibitem{lu2019grid}
Xin Lu, Buyu Li, Yuxin Yue, Quanquan Li, and Junjie Yan.
\newblock Grid r-cnn.
\newblock In {\em CVPR}, 2019.

\bibitem{malisiewicz2011ensemble}
Tomasz Malisiewicz, Abhinav Gupta, and Alexei~A Efros.
\newblock Ensemble of exemplar-svms for object detection and beyond.
\newblock In {\em ICCV}, 2011.

\bibitem{pan2019transferrable}
Yingwei Pan, Ting Yao, Yehao Li, Yu Wang, Chong-Wah Ngo, and Tao Mei.
\newblock Transferrable prototypical networks for unsupervised domain
  adaptation.
\newblock In {\em CVPR}, 2019.

\bibitem{redmon2016you}
Joseph Redmon, Santosh Divvala, Ross Girshick, and Ali Farhadi.
\newblock You only look once: Unified, real-time object detection.
\newblock In {\em CVPR}, 2016.

\bibitem{ren2018learning}
Mengye Ren, Wenyuan Zeng, Bin Yang, and Raquel Urtasun.
\newblock Learning to reweight examples for robust deep learning.
\newblock {\em ICML}, 2018.

\bibitem{ren2015faster}
Shaoqing Ren, Kaiming He, Ross Girshick, and Jian Sun.
\newblock Faster r-cnn: Towards real-time object detection with region proposal
  networks.
\newblock In {\em NeurIPS}, 2015.

\bibitem{shrivastava2016training}
Abhinav Shrivastava, Abhinav Gupta, and Ross Girshick.
\newblock Training region-based object detectors with online hard example
  mining.
\newblock In {\em CVPR}, 2016.

\bibitem{shu2019meta}
Jun Shu, Qi Xie, Lixuan Yi, Qian Zhao, Sanping Zhou, Zongben Xu, and Deyu Meng.
\newblock Meta-weight-net: Learning an explicit mapping for sample weighting.
\newblock {\em NeurIPS}, 2019.

\bibitem{uijlings2013selective}
Jasper~RR Uijlings, Koen~EA Van De~Sande, Theo Gevers, and Arnold~WM Smeulders.
\newblock Selective search for object recognition.
\newblock {\em IJCV}, 2013.

\bibitem{viola2001adaboost}
Paul Viola and Michael Jones.
\newblock Rapid object detection using a boosted cascade of simple features.
\newblock In {\em CVPR}, 2001.

\bibitem{wang2019region}
Jiaqi Wang, Kai Chen, Shuo Yang, Chen~Change Loy, and Dahua Lin.
\newblock Region proposal by guided anchoring.
\newblock In {\em CVPR}, 2019.

\bibitem{wu2019iou}
Shengkai Wu and Xiaoping Li.
\newblock Iou-balanced loss functions for single-stage object detection.
\newblock {\em arXiv preprint arXiv:1908.05641}, 2019.

\bibitem{xie2017aggregated}
Saining Xie, Ross Girshick, Piotr Doll{\'a}r, Zhuowen Tu, and Kaiming He.
\newblock Aggregated residual transformations for deep neural networks.
\newblock In {\em CVPR}, 2017.

\bibitem{yao2015semi}
Ting Yao, Yingwei Pan, Chong-Wah Ngo, Houqiang Li, and Tao Mei.
\newblock Semi-supervised domain adaptation with subspace learning for visual
  recognition.
\newblock In {\em CVPR}, 2015.

\bibitem{zhang2018single}
Shifeng Zhang, Longyin Wen, Xiao Bian, Zhen Lei, and Stan~Z Li.
\newblock Single-shot refinement neural network for object detection.
\newblock In {\em CVPR}, 2018.

\bibitem{zhang2018generalized}
Zhilu Zhang and Mert Sabuncu.
\newblock Generalized cross entropy loss for training deep neural networks with
  noisy labels.
\newblock In {\em NeurIPS}, 2018.

\end{thebibliography}
}

\end{document}